\documentclass[sn-mathphys]{sn-jnl}% Math and Physical Sciences Reference Style
\jyear{2022}%

\theoremstyle{thmstyleone}%
\theoremstyle{thmstyletwo}%

\theoremstyle{thmstylethree}%

\begin{document}

\title[A Prototype-Based Neural Network for Image Anomaly Detection and Localization]{A Prototype-Based Neural Network for Image Anomaly Detection and Localization}


\author[1]{\fnm{Chao} \sur{Huang}}\email{chaohuang@std.uestc.edu.cn}
\author[1]{\fnm{Zhao} \sur{Kang}}\email{zkang@uestc.edu.cn}
\author*[1]{\fnm{Hong} \sur{Wu}}\email{hwu@uestc.edu.cn}

\affil*[1]{\orgdiv{School of Computer Science and Engineering}, \orgname{University of Electronic Science and Technology of China}, \orgaddress{\city{Chengdu}, \postcode{611731}, \state{Sichuan}, \country{China}}}


\abstract{Image anomaly detection and localization perform not only image-level anomaly classification but also locate pixel-level anomaly regions. Recently, it has received much research attention due to its wide application in various fields. This paper proposes ProtoAD, a prototype-based neural network for image anomaly detection and localization. First, the patch features of normal images are extracted by a deep network pre-trained on nature images. Then, the prototypes of the normal patch features are learned by non-parametric clustering. Finally, we construct an image anomaly localization network (ProtoAD) by appending the feature extraction network with $L2$ feature normalization, a $1\times1$ convolutional layer, a channel max-pooling, and a subtraction operation. We use the prototypes as the kernels of the $1\times1$ convolutional layer; therefore, our neural network does not need a training phase and can conduct anomaly detection and localization in an end-to-end manner. Extensive experiments on two challenging industrial anomaly detection datasets, MVTec AD and BTAD, demonstrate that ProtoAD achieves competitive performance compared to the state-of-the-art methods with a higher inference speed. The source code is available at: \url{https://github.com/98chao/ProtoAD}.}

% \abstract{\textbf{Purpose:} The abstract serves both as a general introduction to the topic and as a brief, non-technical summary of the main results and their implications. The abstract must not include subheadings (unless expressly permitted in the journal's Instructions to Authors), equations or citations. As a guide the abstract should not exceed 200 words. Most journals do not set a hard limit however authors are advised to check the author instructions for the journal they are submitting to.
% 

\maketitle
\section{Introduction}\label{sec1}

\begin{figure}[!t]
    \centering
    \includegraphics[width=105mm]{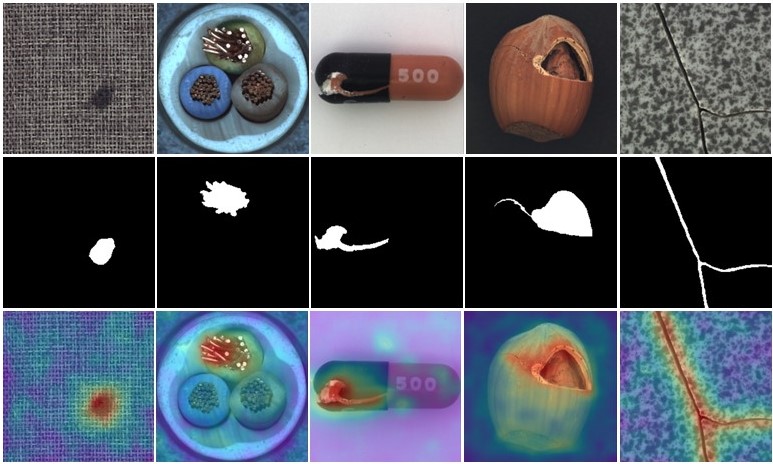}
    \caption{Examples from the MVTec benchmark datasets. From top to bottom: anomaly samples, anomaly mask, and anomaly score maps predicted by our method.}
    \label{fig:ad_examples}
\end{figure}

\textit{Anomaly detection} (AD) \cite{chandola2009anomaly,salehi2022unified} aims to detect anomalous samples that are deviated from a set of normal samples predefined during training. Traditional image anomaly detection adopts a semantic AD setting \cite{ ruff2018deep,sabokrou2018adversarially,golan2018deep,bergman2020classification}, where anomaly samples are from unknown semantic classes different from the one normal samples belong to. Recently, detecting and localizing subtle image anomalies has become an important task in computer vision with various applications, such as anomaly or defect detection in industrial optical inspection \cite{mei2018automatic,bergmann2019mvtec}, anomaly detection and localization in video surveillance \cite{sabokrou2017deep,sabokrou2018deep,sabokrou2018avid}, or anomaly detection in medical images \cite{schlegl2017unsupervised,li2018thoracic}. In this setting, anomaly detection determines whether an image contains any anomaly, and anomaly localization, aka anomaly segmentation, localizes the anomalies at the pixel level. This paper focuses on the second setting, especially industrial anomaly detection and localization. Some examples from the MVTec AD dataset \cite{bergmann2019mvtec} along with predictions by our method are shown in Figure \ref{fig:ad_examples}.

In the above applications, anomalous samples are scarce and hard to collect. Therefore, image anomaly detection and localization are often solved with only normal samples. In addition, anomalous regions within images are often subtle (see Figure \ref{fig:ad_examples}), making image anomaly localization a more challenging task that has not been thoroughly studied compared to image anomaly detection. Recent anomaly localization methods can be roughly categorized into two classes: reconstruction-based methods and OOD-based (out-of-distribution based) methods.

Reconstruction-based methods are mainly based on the assumption that a model trained only on normal images can not reconstruct anomalous images accurately. They reconstruct image as a whole \cite{bergmann2019mvtec,an2015variational,schlegl2017unsupervised, 
bergmann2018improving,gong2019memorizing,liu2020towards,park2020learning,zavrtanik2021reconstruction,hou2021divide,zavrtanik2021draem}, or reconstruct in the feature space ~\cite{bergmann2020uninformed,salehi2021multiresolution,wang2021glancing}. Then anomaly detection and localization can be performed by measuring the difference between the reconstructed and original ones. This kind of method always needs cumbersome network training.

OOD-based methods evaluate the degree of abnormality for a patch feature by measuring its deviation from a set of normal patch features, which is intrinsically a patch-wise OOD detecting task. Some methods such as PatchSVDD \cite{yi2021patch} and CutPaste \cite{li2021cutpaste} learn feature representation by self-supervised learning. On the contrary, some other methods \cite{napoletano2018anomaly,cohen2020sub,defard2021padim,roth2022towards} simply extract features by deep networks pre-trained on natural image datasets such as ImageNet\cite{deng2009imagenet}, and achieve promising and even better performances. Since the number of training patches is much larger than that of training images, the inference time and storage increase remarkably. Different strategies have been proposed to tackle this problem. Napoletano et al. \cite{napoletano2018anomaly} used k-means to learn the dictionary/prototypes for normal patch features, but they evaluated each test patch independently, resulting in high inference time. SPADE \cite{cohen2020sub} selects k-nearest normal images for patch-wise evaluation based on the global image features, limiting anomaly localization performance. PaDiM \cite{defard2021padim} models the normal patches at each position by a multidimensional Gaussian distribution and measures the anomaly by the Mahalanobis distance between a test patch feature and the Gaussian at the same position. However, both SPADE \cite{cohen2020sub} and PaDiM \cite{defard2021padim} are reliant on image alignment. The current state-of-the-art method, PatchCore \cite{roth2022towards}, uses greedy coreset subsampling to reduce the inference time and storage significantly.

This paper proposes ProtoAD, a prototype-based neural network for image anomaly detection and localization, to improve OOD-based methods' inference speed. We assume that all normal patch features can be grouped into some prototypes, and abnormal patch features cannot be properly assigned to any of them. Therefore, image anomaly localization can be performed by measuring the deviation of test patch features from the prototypes of normal patch features. First, the patch features of normal images are extracted by a deep network pre-trained on nature images and are $L2$-normalized. Then the prototypes of the normalized normal patch features are learned by a non-parametric clustering algorithm. The cosine similarity between two $L2$-normalized vectors is equivalent to the dot product between them. Therefore the cosine similarity between a normalized patch feature and a prototype can be implemented by a $1\times1$ convolution. Based on this equivalence, we construct an image anomaly localization network (ProtoAD) by appending the feature extraction network with the $L2$ feature normalization, a $1\times1$ convolutional layer, a channel max-pooling, and a subtraction operation. We use the prototypes as the kernels of the $1\times1$ convolutional layer; therefore, our neural network does not need a training phase. Compared with previous OOD-based methods \cite{napoletano2018anomaly,cohen2020sub,defard2021padim,roth2022towards}, ProtoAD can perform the anomaly detection and localization in an end-to-end manner, which is more elegant and efficient. Extensive experiments on two challenging industrial anomaly detection datasets, MVTec AD \cite{bergmann2019mvtec} and BTAD \cite{mishra2021vt}, demonstrate that ProtoAD achieves competitive performance compared to the state-of-the-art methods with a higher inference speed. This advantage of ProtoAD makes it better match the needs of real-world industrial applications.

\section{Related Works}
\label{sec:related_work}

\subsection{Image Anomaly Localization}
Anomaly detection is an image-level task to determine whether an image contains any anomaly. On the other hand, anomaly localization is more complex to locate anomalies at the pixel level. Here, we only introduce the methods that can be directly applied to image anomaly localization and roughly categorize current methods into two types: reconstruction-based and OOD-based. 

Reconstruction-based methods are mainly based on the assumption that a model trained only on normal images can not reconstruct anomalous images accurately, and anomaly detection and localization can be performed by measuring the difference between the reconstructed and original images. Early reconstruction-based methods \cite{bergmann2019mvtec,an2015variational,schlegl2017unsupervised,
bergmann2018improving,liu2020towards} reconstruct image by auto-encoders (AE), variational autoencoders (VAE) or generative adversarial networks (GAN). However, the neural networks have high generalization capacities and can reconstruct anomalies well. Later, different strategies have been proposed to tackle this problem. Different memory-based auto-encoders \cite{gong2019memorizing,park2020learning,hou2021divide} have been proposed to reconstruct images with features from memory bank to limit the generalization ability. Student-teacher models \cite{bergmann2020uninformed,salehi2021multiresolution} have been used to reconstruct pre-trained deep features. RIAD \cite{zavrtanik2021reconstruction} randomly removes partial image regions and reconstructs the image by image in-painting. Glance \cite{wang2021glancing} trains a Global-Net to regress the deep features of cropped patches based on their context. DRAEM \cite{zavrtanik2021draem} combines a reconstructive sub-network and a discriminative network and trains them in an end-to-end manner on synthetically generated just-out-of-distribution images. 

OOD-based methods evaluate the degree of abnormality for a patch feature by measuring its deviation from a set of normal patch features, which is intrinsically a patch-wise OOD detecting task. Some methods such as PatchSVDD \cite{yi2021patch} and CutPaste \cite{li2021cutpaste} learn feature representation by self-supervised learning. On the contrary, some other methods \cite{napoletano2018anomaly,cohen2020sub,defard2021padim,roth2022towards} simply extract features by deep networks pre-trained on natural image datasets such as ImageNet \cite{deng2009imagenet}, and achieve promising and even better performances. Since the number of training patches is much larger than that of training images, the inference time and storage increase remarkably. Different strategies such as clustering, density estimation, and sampling have been proposed to tackle this problem. Napoletano et al. \cite{napoletano2018anomaly} learned a dictionary of normal patches from the training set by k-means, and evaluated each patch of a test image by measuring its visual similarity with the k-nearest neighbors in the dictionary. SPADE \cite{cohen2020sub} compares patch features of a test image with the patch features at the same position of k-nearest normal images selected based on global image features. However, this oversimplified pre-selection strategy will limit the localization performance. PaDiM \cite{defard2021padim} models the normal patches at each position by a multidimensional Gaussian distribution and detect anomaly by the Mahalanobis distance between a test patch feature and the Gaussian at the same position. Both SPADE \cite{cohen2020sub} and PaDiM \cite{defard2021padim} are reliant on image alignment. Recently, PatchCore \cite{roth2022towards} constructs the memory bank of locally aware patch features by greedy coreset subsampling, and localizes anomaly by measuring the distances of test patch features to their nearest normal patch features in the bank. As a result, PatchCore achieves a new state-of-the-art and significantly reduces the inference time and storage.

Our method is also an OOD-based method with pre-trained deep features but has several differences from the previous works. Our method uses non-parametric clustering instead of k-means in \cite{napoletano2018anomaly} to learn the prototypes for normal patch features. More importantly, our method can perform anomaly detection and localization by a network in an end-to-end manner, which is more elegant and efficient than the previous methods. Compared to reconstruction-based methods, our network do not need a cumbersome network training phase.

\subsection{Clustering Algorithms}
Clustering is a type of unsupervised learning task of dividing a set of unlabeled data points into a number of groups such that the data points in the same groups are more similar to each other than they are to the data points in other groups. Clustering provides an abstraction from data points to the clusters, and each cluster can be characterized by a cluster prototype, such as the centroid of a cluster, for further analysis. Clustering algorithms can be roughly divided into four categories: Partition-based cluster, Density-based clustering, Spectral Clustering, and Hierarchical-based clustering.

Partition-based clustering algorithms divide the data into k groups, where k is the predefined number of cluster. The classical algorithms are k–means \cite{1965Some} and its variations. Although these algorithms are very fast, they need the number of clusters as a parameter and are sensitive to the selection of the initial k centroids.

Density-based clustering defines a cluster as the largest set of densely connected points and can find clusters of arbitrary shapes. DBSCAN \cite{ester1996density} is the most representative algorithm of this class. It has two parameters, radius length $\epsilon$ and a parameter $MinPts$. If there are $MinPts$ points in the radius of $\epsilon$ of a point, it is regarded as a high-density point.

Spectral Clustering \cite{von2007tutorial} has recently attracted much attention. Most spectral clustering algorithms need to compute the full similarity graph Laplacian matrix and have quadratic complexities, thus severely restricting their application to large data sets.

Hierarchical clustering \cite{karypis1999chameleon} is of two types: bottom-up and top-down approaches. In the bottom-up approach (aka agglomerative clustering), each data point starts as a cluster, and the most similar cluster pairs are iteratively merged according to the chosen similarity measure until some stopping criteria are met. In the top-down approach (aka divisive clustering), the clustering begins with a large cluster including all data and recursively breaks down into smaller clusters. Hierarchical clustering produces a clustering tree that provides meaningful ways to interpret data at different levels of granularity. Recently, Sarfraz et al. \cite{sarfraz2019efficient} proposed FINCH, a high-speed, scalable, and fully parameter-free hierarchical agglomerative clustering algorithm.

In \cite{napoletano2018anomaly}, k-means is used to learn the prototypes from normal patch features. To avoid choosing the number of clusters ahead, we adopt FINCH to learn the prototypes for normal patch features.

\begin{figure}[!t]
\centering
\includegraphics[width=105mm]{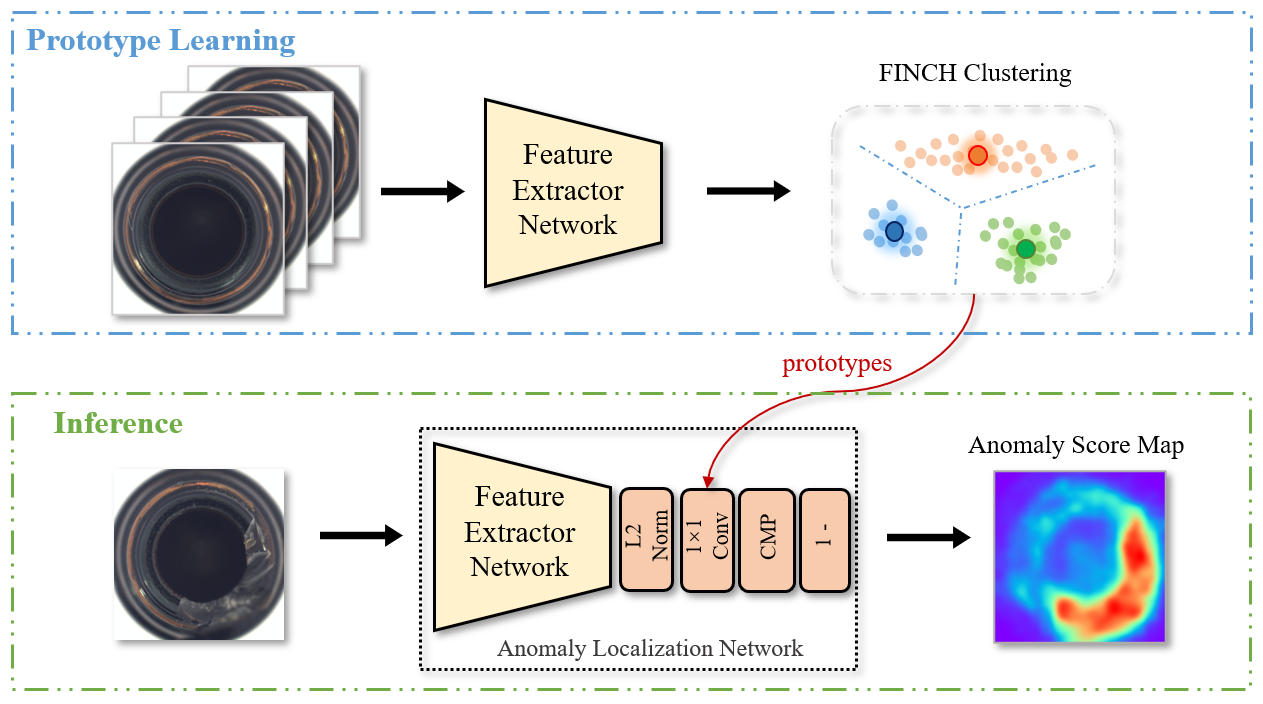} 
\caption{
An overview of the proposed method. First, the patch features of normal images are extracted by a deep network pre-trained on nature images. Then, the prototypes of the normal patch features are learned by FINCH clustering. For inference, an image anomaly localization network (ProtoAD) is constructed by appending the feature extraction network with the $L2$ feature normalization, a $1\times1$ convolutional layer, a channel max-pooling (CMP), and a subtraction operation, and anomaly localization is performed in an end-to-end manner.}
\label{pipeline}
\end{figure}

\section{Method}
\label{method}
Our method consists of three steps: patch feature extraction, prototype learning, and anomaly detection and localization. An overview of our method is given in Figure. \ref{pipeline}. We describe them sequentially in the following subsection.

\subsection{Patch Feature Extraction}

Since the features extracted by pre-trained networks have shown their effectiveness for various visual applications including anomaly detection \cite{napoletano2018anomaly,cohen2020sub,defard2021padim,roth2022towards,bergmann2020uninformed,salehi2021multiresolution}, we also adopt deep networks pre-trained on ImageNet dataset \cite{deng2009imagenet} as the feature extractor, and choose the backbone of Wide-ResNet \cite{zagoruyko2016wide} as the feature extractor following the previous works \cite{cohen2020sub,defard2021padim,roth2022towards}.

ResNet-like deep networks \cite{he2016deep,zagoruyko2016wide} include several convolutional stages. The features become more abstract when the stage goes deeper, but their resolution gets lower. Thus, the feature maps from different stages form a feature hierarchy for an input image. Each spatial position of a feature map has a receptive field and corresponds to a patch/region in an input image; therefore, the feature vector at a spatial position of feature maps can be considered as a feature representation for the corresponding image patch. If the feature maps of a stage have a resolution of $H\times W$, they contains $H\times W$ patch features. The deep and abstract features from the ImageNet pre-trained networks are biased towards the ImageNet classification task and are less relevant to the anomaly detection and localization task. Therefore, we adopt the low- and mid-level (stage 1-3) feature representations and combine them as the patch features. Concretely, the feature maps at the higher-level are bilinearly re-scaled to have the same resolution as the lowest level, then the feature maps at different levels are concatenated together for handling multi-scale anomalies. The extracted features are then $L2$-normalized where each feature vector is divided by its $L2$ norm.

\subsection{Prototype Learning}

After feature extraction, the prototypes of the $L2$-normalized patch features are learned by a clustering algorithm. Then, the prototypes are used in anomaly detection and localization instead of all the normal patch features to reduce the inference time and storage. There are mainly two concerns in choosing a clustering algorithm. First, the number of patch features is much larger than that of training images. For example, each category of MVTec AD dataset has several hundreds of images, while it has several hundreds of thousands of patch features in our implementation. Therefore, the clustering algorithm should be efficient and scalable to large-scale data. Second, most clustering algorithms have some parameters, e.g., the number of clusters or distance thresholds, which can not be well set without a priori knowledge of the data distribution. Thus, these algorithms demand a tedious parameter tuning process to achieve good performance. To meet the requirements of real applications, we adopt FINCH \cite{sarfraz2019efficient}, a high-speed, scalable, and fully parameter-free hierarchical agglomerative clustering algorithm. 

The core idea of FINCH is to use the nearest neighbor information of each data point for clustering, which does not need to specify any parameters and has a low computational overhead. Given the integer indices of the first neighbor of each data point, an adjacency matrix is defined according to the following rules:
\begin{equation}
    A(i,j)=\left\{
        \begin{aligned}
        1,\text{   } & \text{ if }j=\kappa_i^1 \text{ or } \kappa_j^1=i \text{ or } \kappa_i^1=\kappa_j^1 \\
        0,\text{   } & otherwise
        \end{aligned}
        \right.
\end{equation}
where $\kappa_i^1$ symbolizes the first neighbor of data point $i$. This sparse adjacency matrix specifies a graph where connected data points form clusters. It directly provides clusters without solving a graph segmentation problem. After computing the first partition, FINCH merges the clusters recursively by using cluster means to compute the first neighbor of each cluster until all data points are included in a single cluster or until some stopping criteria is met. In this work, we define the stopping criteria as the number of cluster is less than a threshold and set the threshold to 10,000 to get good results in our experiments. We choose the last partition as the clustering result, and use the mean vectors of clusters as the prototypes of normal patch features.

When the features are $L2$-normalized (making the length of a vector to 1), cosine similarity and Euclidean distance between the normalized features are equivalent in the sense of nearest neighbor searching:
\begin{equation}
    \begin{split}
      \frac{1}{2}{L_2(\mathbf{x}_a,\mathbf{x}_b)}^2  = \frac{1}{2}(\mathbf{x}_a-\mathbf{x}_b)\cdot{(\mathbf{x}_a-\mathbf{x}_b)} = 1-\mathbf{x}_a\cdot \mathbf{x}_b=1-\cos{(\mathbf{x}_a,\mathbf{x}_b)} \\
    \end{split}
\end{equation}
where $L_2()$ is Euclidean distance, $\mathbf{x}_a$ and $\mathbf{x}_b$ are two $L2$-normalized feature vectors, and $\cos$ is cosine similarity. Therefore, we use cosine similarity for clustering and measuring the deviation of test patch features from norm patch features in the next subsection.

\subsection{Neural Network for Anomaly Detection and Localization}

When a test image passes through the feature extraction network, $H\times W$ patch features have been extracted. The anomaly score of each patch feature can be computed by measuring its deviation from the prototypes of normal patch features. We compute the anomaly score of a test patch as one minus the cosine similarity between the normalized test patch feature and its nearest prototype. Formally, the anomaly score for the patch at position $(i,j)$ can be calculated as
\begin{equation}
s_{ij} = 1 - \max\limits_{1\leqslant k \leqslant K}\cos{(\mathbf{x}_{ij}, \mathbf{m}_k)}
\end{equation}
where $\mathbf{x}_{ij}$ is the normalized patch feature at position ($i,j$), $\mathbf{m}_k$ is the $k$-th prototype, and $\cos$ is cosine similarity. In addition, the image-level anomaly score for a test image can be simply computed by maximizing the anomaly scores of all its patch features.
\begin{equation}
S = \max \limits_{1\leqslant i \leqslant H, 1\leqslant j \leqslant W} s_{ij} 
\end{equation}

\begin{figure}[!t]
\centering
\includegraphics[width=105mm]{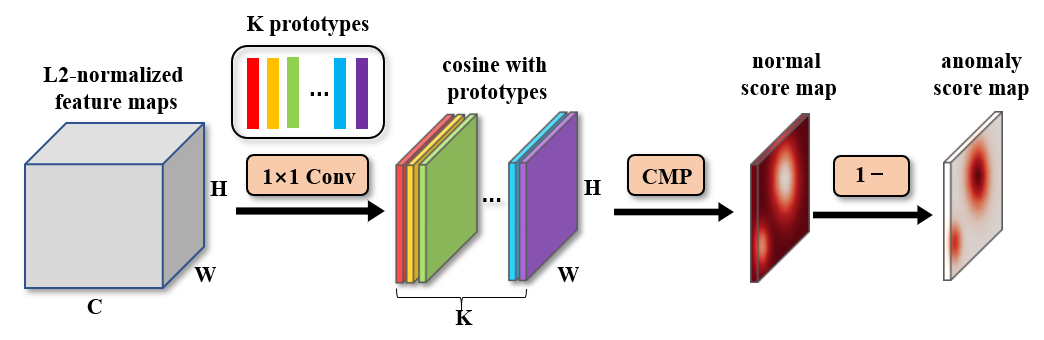}
\caption{Anomaly detection and localization process of ProtoAD.}
\label{process}
\end{figure}

The cosine similarities between a normalized patch feature and a prototype can be computed by a $1\times1$ convolution (dot product) between them. Based on this equivalence, we construct a neural network (ProtoAD) for anomaly detection and localization. First, the $L2$ feature normalization and a $1\times1$ convolutional layer are appended to the feature extraction network, and outputs feature maps of size $H\times W \times K$, including the cosine similarities between the $H\times W$ normalized patch features and all $K$ prototypes. Then, channel max-pooling (CMP) is applied to the feature maps to get the normal score map of $H\times W$, including the cosine similarities between the $H\times W$ normalized patch features and their nearest prototypes. The anomaly score map can be further obtained by computing one minus the normal score map. This process is illustrated by Figure \ref{process}. Since the spatial resolution of feature maps is lower than that of an input image, we resize the anomaly score map to the resolution of the input image and use a Gaussian filter to smooth it. Finally, anomaly localization can be achieved by thresholding the anomaly score map, and the anomaly score for the test image can be obtained by maximizing the anomaly score map. 

We use the prototypes of normal patch features as the kernels of the $1\times1$ convolutional layer. Therefore the proposed neural network does not need a training phase. Compared to previous works \cite{napoletano2018anomaly,cohen2020sub,defard2021padim,roth2022towards}, our method can perform the anomaly detection and localization in an end-to-end manner, which is more elegant and efficient.

\section{Experiments}

%------------------------------------------------------------------------
\subsection{Datasets and Metrics}

\subsubsection{Dataset} MVTec AD dataset \cite{bergmann2019mvtec} is a real-world industrial defect detection dataset which has become a standard benchmark for evaluating image anomaly detection and localization methods. It has 5354 high-resolution images belonging to 10 objects and 5 texture categories. The images of each category are split into a training and a testing set. Totally, the training set has 3629 normal images, and the test set has 1725 normal and abnormal images of various defects. The ground truth of the test set contains anomaly labels for image-level evaluation and anomaly masks for pixel-level evaluation. 

BTAD (BeanTech Anomaly Detection dataset) is a real-world industrial dataset recently released by \cite{mishra2021vt}. It contains a total of 2830 real-world images of 3 industrial products. The images of each category are split into a defect-free training set and a testing set, supporting evaluation of both anomaly detection and localization.
 
We follow the split of the two datasets for training and testing.

\subsubsection{Evaluation Metrics} AUROC (Area Under the Receiver Operating Characteristic curve) is the most commonly used metric for anomaly detection, which is independent of the threshold. We use image-level AUROC for evaluating the performance of anomaly detection, pixel-level AUROC for anomaly localization. Since the pixel-level AUROC is biased in favor of large anomalies, we also use PRO-score (per-region-overlap) \cite{bergmann2020uninformed} to evaluate anomaly localization, which weights ground-truth regions of different sizes equally.

\subsection{Experimental Setup}
We normalize the size of images from all categories of MVTec AD and BTAD dataset to $256\times 256$, center crop images to $224\times 224$, and do not apply any data augmentation. The backbone of Wide-ResNet50 pre-trained on ImageNet is employed as the feature extractor in our method as in \cite{cohen2020sub,defard2021padim,roth2022towards}. We define the stopping criteria for FINCH clustering algorithm as the number of clusters is less than 10,000 and choose the last generated partition as the clustering result. For inference, we up-sample the anomaly score map to image size using bilinear interpolation and smooth it with the Gaussian filter with parameter $\delta=4$ as in \cite{defard2021padim}. We implemented our models in Python 3.7 \cite{van2009python} and PyTorch \cite{paszke2019pytorch}, and run experiments on NVIDIA GeForce RTX 2080 Ti.

\subsection{Results on MVTec AD}
\subsubsection{Comparison with the State-of-the-art} 
We compare ProtoAD with the state-of-the-art methods including both the reconstruction and OOD-based methods. The compared  reconstruction-based methods include Uninformed students (U-Student) \cite{bergmann2020uninformed}, RIAD \cite{zavrtanik2021reconstruction}, MKD \cite{salehi2021multiresolution}, Glance \cite{wang2021glancing}, DAAD \cite{hou2021divide} and DREAM \cite{zavrtanik2021draem}. And the compared OOD-based methods include SPADE \cite{cohen2020sub}, PatchSVDD (P-SVDD) \cite{yi2021patch} , CutPaste \cite{li2021cutpaste}, PaDiM \cite{defard2021padim}, and PatchCore (P-Core) \cite{roth2022towards}. We directly use their evaluation results if they have been provided.

\begin{table}[ht]
    \centering
    \caption{Anomaly localization performance on MVTec AD (Pixel-level AUROC). The best results of the two classes of methods are bold-faced respectively.}
    \label{tab:mvtec_segmentation}
    \resizebox{\textwidth}{!}{
    \begin{tabular}{l c c c c c c c c c c}
        \hline
        \multirow{2}{*}{\bf Category} & \multicolumn{4}{c}{\bf Reconstruction-based} & \multicolumn{6}{c}{\bf OOD-based}  \\
        \cline{2-11}
        & \bf MKD	& \bf Glance	& \bf RIAD	& \bf DRAEM	& \bf P-SVDD	& \bf CutPaste	& \bf SPADE	& \bf PaDiM	& \bf P-Core & \bf ProtoAD \\ \hline
        Carpet	    & 95.6 &	96.0 &	96.3 & 	95.5 &	92.6 &	98.3 &	97.5 &	99.1 &	99.0 &	99.2 \\
        Grid	    & 91.8 &	78.0 &	98.8 & 	99.7 &	96.2 &	97.5 &	93.7 &	97.3 &	98.7 &	98.0 \\
        Leather	    & 98.1 &	90.0 &	99.4 & 	98.6 &	97.4 &	99.5 &	97.6 &	99.2 &	99.3 &	99.4 \\
        Tile	    & 82.8 &	80.0 &	89.1 & 	99.2 &	91.4 &	90.5 &	87.4 &	94.1 &	95.6 &	95.2 \\
        Wood	    & 84.8 &	81.0 &	85.8 & 	96.4 &	90.8 &	95.5 &	88.5 &	94.9 &	95.0 &	95.6 \\ \hline
        \textbf{Texture}	    & 90.6 &	85.0 &	93.9 & 	97.9 &	93.7 &	96.3 &	92.9 &	96.9 &	97.5 &	97.5 \\ \hline
        Bottle	    & 96.3 &	93.0 &	98.4 & 	99.1 &	98.1 &	97.6 &	98.4 &	98.3 &	98.6 &	98.3 \\
        Cable	    & 82.4 &	94.0 &	84.2 & 	94.7 &	96.8 &	90.0 &	97.2 &	96.7 &	98.4 &	97.5 \\
        Capsule	    & 95.9 &	90.0 &	92.8 & 	94.3 &	95.8 &	97.4 &	99.0 &	98.5 &	98.8 &	98.2 \\
        Hazelnut	& 94.6 &	84.0 &	96.1 & 	99.7 &	97.5 &	97.3 &	99.1 &	98.2 &	98.7 &	98.8 \\
        Metal Nut	& 86.4 &	91.0 &	92.5 & 	99.5 &	98.0 &	93.1 &	98.1 &	97.2 &	98.4 &	96.8 \\
        Pill	    & 89.6 &	93.0 &	95.7 & 	97.6 &	95.1 &	95.7 &	96.5 &	95.7 &	97.4 &	94.2 \\
        Screw	    & 96.0 &	96.0 &	98.8 & 	97.6 &	95.7 &	96.7 &	98.9 &	98.5 &	99.4 &	98.9 \\
        Toothbrush	& 96.1 &	96.0 &	98.9 & 	98.1 &	98.1 &	98.1 &	97.9 &	98.8 &	98.7 &	98.8 \\
        Transistor	& 76.5 &	100  & 	87.7 & 	90.9 &	97.0 &	93.0 &	94.1 &	97.5 &	96.3 &	92.5 \\
        Zipper	    & 93.9 &	99.0 &	97.8 & 	98.8 &	95.1 &	99.3 &	96.5 &	98.5 &	98.8 &	96.7 \\ \hline
        \textbf{Object}       & 90.8 &	93.6 &	94.3 & 	97.0 &	96.7 &	95.8 &	97.6 &	97.8 &	98.4 &	97.1 \\ \hline
        \textbf{All}      & 90.7 &	90.7 &	94.2 & 	\textbf{97.3} &	95.7 &	96.0 &	96.0 &	97.5 &	\textbf{98.1} &	97.2 \\ \hline
        \end{tabular}}
\end{table}

\begin{table}[ht]
    \centering
    \caption{Anomaly localization performance on MVTec AD (PRO-score). The best results of the two classes of methods are bold-faced respectively.}
    \label{tab:mvtec_segmentation_pro}
    \resizebox{0.75\textwidth}{!}{
    \begin{tabular}{l c c c c c c}     			
        \hline
        \multirow{2}{*}{\bf Category} & \multicolumn{2}{c}{\bf Reconstruction-based} & \multicolumn{4}{c}{\bf OOD-based}  \\
        \cline{2-7}
         & \bf U-Student & \bf Glance & \bf SPADE & \bf PaDiM & \bf P-Core & \bf ProtoAD\\ \hline
        Carpet	    & 87.9 & 	97.7 &	94.7 	& 96.2 &	96.6 &	97.0  \\ 
        Grid	    & 95.2 & 	93.2 &	86.7 	& 94.6 &	96.0 &	93.9  \\
        Leather	    & 94.5 & 	90.9 &	97.2 	& 97.8 &	98.9 &	98.1  \\
        Tile	    & 94.6 & 	88.3 &	75.6 	& 86.0 &	87.3 &	87.2  \\
        Wood	    & 91.1 & 	94.1 &	87.4 	& 91.1 &	89.4 &	93.2  \\ \hline
        \textbf{Texture} & 92.7 & 	92.8 &	88.3 	& 93.1 &	93.6 &	93.9 \\ \hline
        Bottle  	& 93.1 & 	96.8 &	95.5 	& 94.8 &	96.2 &	95.8  \\
        Cable   	& 81.8 & 	98.0 &	90.9 	& 88.8 &	92.5 &	93.8  \\
        Capsule	    & 96.8 & 	96.0 &	93.7 	& 93.5 &	95.5 &	93.7  \\
        Hazelnut	& 96.5 & 	96.2 &	95.4 	& 92.6 &	93.8 &	95.3  \\
        Metal Nut	& 94.2 & 	96.7 &	94.4 	& 85.6 &	91.4 &	94.2  \\
        Pill    	& 96.1 & 	97.8 &	94.6 	& 92.7 &	93.2 &	94.7  \\
        Screw   	& 94.2 & 	100  & 	96.0 	& 94.4 &	97.9 &	94.7  \\
        Toothbrush	& 93.3 & 	96.1 &	93.5 	& 93.0 &	91.5 &	91.2  \\
        Transistor	& 66.6 & 	99.9 &	87.4 	& 84.5 &	83.7 &	87.9  \\
        Zipper	    & 95.1 & 	99.2 &	92.6 	& 95.9 &	97.1 &	93.3  \\
        \textbf{Object}	& 90.8 & 	97.7 &	93.4 	& 91.6 &	93.3 &	93.4 \\   \hline
        \textbf{All}	& 91.4 & 	\textbf{96.1} &	91.7 	& 92.1 &	93.4 &	\textbf{93.6} \\   
        \hline 
    \end{tabular}}
\end{table}

We report the evaluation results (pixel-level AUROC and PRO-score) for pixel-level anomaly localization on MVTec AD dataset in Table \ref{tab:mvtec_segmentation} and Table \ref{tab:mvtec_segmentation_pro} respectively. From Table \ref{tab:mvtec_segmentation}, we can see that the OOD-based methods generally achieve better pixel-level AUROC than the reconstruct-based methods. Among the OOD-based methods, the methods using the pre-trained deep features achieve better pixel-level AUROC than the methods based on self-supervised learning. PatchCore achieves the best pixel-level AUROC, PaDiM the second, and the reconstruct-based method DREAM the third. The pixel-level AUROC of our method is very close to those of PaDiM and DREAM. We also notice that our method is more effective on the texture category and achieves the second best AUROC. Table \ref{tab:mvtec_segmentation_pro} gives the PRO-score results for methods which have used this metric. Among them, Glance achieves the best result, our method is the second best and outperform other OOD-based methods. After all, our method achieves competitive anomaly localization performance to the state-of-the-art methods.

\begin{figure}[!t]
\centering
\includegraphics[width=110mm]{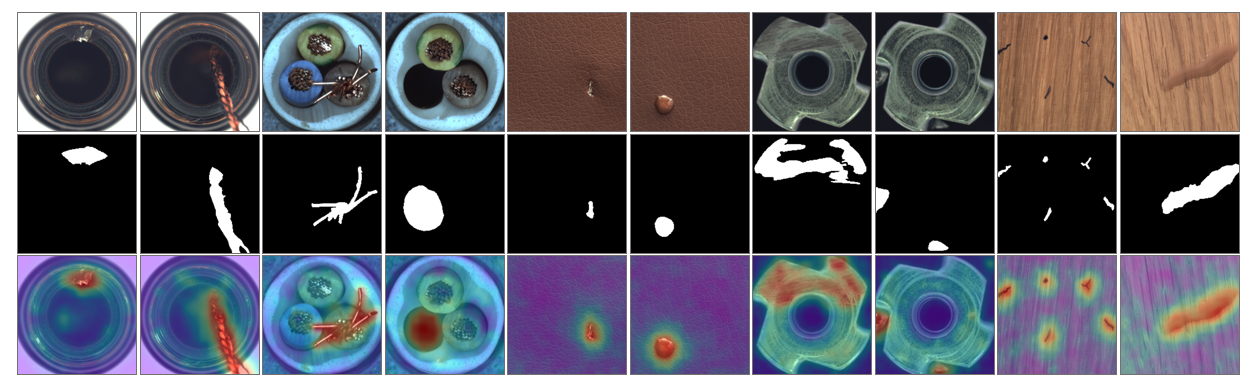} 
\caption{
Qualitative anomaly localization results of our method. From top to bottom: abnormal images, ground-truth, and anomaly score maps produced by our method.
}
\label{visualization}
\end{figure}

Figure \ref{visualization} gives qualitative anomaly localization results of our method on MVTec AD dataset. We can see that our method can give accurate pixel-level localization regardless of anomaly region size and type (see supplementary for more qualitative results). 

We also report the image-level AUROC results for anomaly detection in Table \ref{tab:mvtec_detection}. PatchCore achieves the best AUROC again, DREAM the second. Our method remains competitive and achieves the third-best AUROC, which is very close to that of DREAM. 

\begin{table*}[t]
    \centering
    \caption{Anomaly detection performance on MVTec AD (Image-level AUROC). The best results of the two classes of methods are bold-faced respectively.}
    \label{tab:mvtec_detection}
    \resizebox{\textwidth}{!}{
    \begin{tabular}{l c c c c c c c c c c}
        \hline
        \multirow{2}{*}{\bf Category} & \multicolumn{4}{c}{\bf Reconstruction-based} & \multicolumn{6}{c}{\bf OOD-based}  \\
        \cline{2-11}
         & \bf MKD	 & \bf DAAD 	 & \bf RIAD	 & \bf DRAEM  & \bf SPADE & \bf PaDiM & \bf P-SVDD & \bf CutPaste & \bf P-Core & \bf ProtoAD \\ \hline
        Carpet	    & 79.3 &	86.6 &	84.2 	& 97.0  & 	- 	&	-       & 92.6 & 98.3 &	98.7 &	99.5  \\
        Grid	    & 78.0 &	95.7 &	99.6 	& 99.9  & 	-	&	-       & 96.2 & 97.5 &	98.2 &	93.7  \\
        Leather	    & 95.1 &	86.2 &	100 	& 100   &	-	&	-       & 97.4 & 99.5 &	100	 &  100 \\
        Tile	    & 91.6 &	88.2 &	98.7 	& 99.6  & 	-	&	-       & 91.4 & 90.5 &	98.7 &	99.2  \\
        Wood	    & 94.3 &	98.2 &	93.0 	& 99.1  & 	-	&	-       & 90.8 & 95.5 &	99.2 &	99.1  \\ \hline
        \textbf{Texture}	    & 87.6 &	91.0 &	95.1 	& 99.1  & 	-	 & 98.8 	& 93.7 & 96.3 &	99.0 & 	98.3  \\    \hline
        Bottle	    & 99.4 &	97.6 &	99.9 	& 99.2  & 	-	&	-       & 98.1 & 97.6 &	100	 &  99.9  \\
        Cable   	& 89.2 &	84.4 &	81.9 	& 91.8  & 	-	&	-       & 96.8 & 90.0 &	99.5 &	98.3  \\
        Capsule	    & 80.5 &	76.7 &	88.4 	& 98.5  & 	-	&	-       & 95.8 & 97.4 &	98.1 &	93.1  \\
        Hazelnut	& 98.4 &	92.1 &	83.3 	& 100   &	-	&	-       & 97.5 & 97.3 &	100	 &  100 \\
        Metal Nut	& 73.6 &	75.8 &	88.5 	& 98.7  & 	-	&	-       & 98.0 & 93.1 &	100	 &  99.9  \\
        Pill	    & 82.7 &	90.0 &	83.8 	& 98.9  & 	-	&	-       & 95.1 & 95.7 &	96.6 &	95.8  \\
        Screw	    & 83.3 &	98.7 &	84.5 	& 93.9  & 	-	&	-       & 95.7 & 96.7 &	98.1 &	94.9  \\
        Toothbrush	& 92.2 &	99.2 &	100 	& 100   &	-	&	-       & 98.1 & 98.1 &	100	 &  99.7  \\
        Transistor	& 85.6 &	87.6 &	90.9 	& 93.1  &	-	&	-       & 97.0 & 93.0 &	100	 &  97.7  \\
        Zipper	    & 93.2 &	85.9 &	98.1 	& 100   &	-	&	-       & 95.1 & 99.3 &	99.4 &	94.6  \\    \hline
        \textbf{Object}	        & 87.8 &	88.8 &	89.9 	& 97.4  & 	-	 &  93.6 	& 96.7 & 95.8 &	99.2 & 	97.4  \\    \hline
        \textbf{All} 	        & 87.7 &	89.5 &	91.7 	& \textbf{98.0}  & 	85.5 &	95.3 	& 95.7 & 96.0 &	\textbf{99.1} & 	97.7  \\    \hline
    \end{tabular}}
\end{table*}

\begin{table}[b]
\begin{center}
\caption{Comparison of inference speed. Scores includes image-level AUROC, pixel-level AUROC, and PRO-score. The best results are bold-faced.}
\label{tab:inference_time}
\setlength{\tabcolsep}{4mm}{
\begin{tabular}{ccc}
    \hline
    \noalign{\smallskip}
     Method & $\qquad $ Scores $\qquad\qquad$ & Inference Speed (FPS)\\
    \hline
    SPADE   & (85.5, 96.0, 91.7)  & 7.58 \\
    PatchCore(25\%) & (\textbf{99.1}, \textbf{98.1}, 93.4) & 22.30 \\
    PatchCore(10\%) & (99.0, \textbf{98.1}, 93.5) & 24.45 \\
    PatchCore(1\%)  & (99.0, 98.0, 93.1) & 26.17 \\
    % PaDiM   & (95.3, 97.5, 92.1) & 3.98 \\
    PaDiM  & (95.3, 97.5, 92.1) & 60.32 \\
    ProtoAD    & (97.7, 97.2, \textbf{93.6}) & \textbf{72.45} \\
    \hline
\end{tabular}}
\end{center}
\end{table}

\subsubsection{Inference Efficiency}
Anomaly detection and localization algorithms need high precision and inference speed to match the requirements of real-world applications. Thus, we also report the inference speed of our method and previous OOD-based methods using pre-trained deep features \cite{cohen2020sub,defard2021padim,roth2022towards} in the Table \ref{tab:inference_time}. In the experiments, all the methods adopt Wide-ResNet50 pre-trained on ImageNet as the feature extractor, center-cropped $224\times 224$ image as input, and run on the same machine with a NVIDIA GeForce RTX 2080 Ti. For PatchCore, we use the implementation provided by the authors, which downsamples the normal patch features via greedy coreset subsampling (PatchCore-$x\%$ denotes the percentage x of normal patch features are used in inference) and uses \textit{faiss} \cite{johnson2019billion} for nearest neighbor retrieval and distance computations. For PaDiM, we make extensive optimization via GPU acceleration. Compared with the previous methods, our model achieves the highest speed, which is 1.2x, 2.7x, and 9.5x faster than PaDiM, PatchCore, and SPADE, respectively. The high inference speed is mainly because our model performs inference in an end-to-end manner, and the main computation added to the feature extraction network is the $1\times1$ convolutional layer. Compared to the reconstruct-based methods, our method does not need a cumbersome network training process.

\subsection{Ablation Study}

We report ablations studies on the MVTec AD dataset to evaluate the impact of different components of our method on the performance.

\begin{table}[h]
    \centering
    \caption{Anomaly detection and localization performance of ProtoAD with features at different levels. Each tuple shows image-level AUROC and pixel-level AUROC.}
    \label{tab:feature_layers}
    % \resizebox{0.65\linewidth}{!}{
    \setlength{\tabcolsep}{4.5mm}{
    \begin{tabular}{c c c c}
        \hline
        Feature Level & \quad Texture \quad & \quad Object \quad & All \\ \hline
        level 1  &   (96.2, 96.6) & (85.5, 92.7) & (89.0, 94.0)  \\     
        level 2  &   (97.9, 97.3) & (97.2, 95.5) & (97.4, 96.1) \\      
        level 3  &   (98.0, 96.7) & (95.7, 96.0) & (96.5, 96.2) \\      
        level 2+3    &  (97.9, 97.3) & (96.8, 96.9) & (97.1, 96.9) \\   
        level 1+2+3  &  (98.3, 97.5) & (97.4, 97.1) & (97.7, 97.2)  \\  \hline
    \end{tabular}}
\end{table}

\subsubsection{Feature Layer Selection}

\label{ablation:feature_layers}
ResNet-like deep networks \cite{he2016deep,zagoruyko2016wide} include several convolutional stages. The feature maps from different stages can compose a feature hierarchy for an image. Since the deepest feature maps in the hierarchy are biased towards the ImageNet classification task, we only adopt the features at the low and middle hierarchy levels (stage 1-3) for anomaly detection and localization. Table \ref{tab:feature_layers} gives the performance achieved with the features from different levels and their combination. It can be observed that the features from hierarchy level 2 can achieve the best performance among the first three levels, and a combination of the three levels can further improve the performance. Therefore, our method uses the combination of the first three feature levels as the patch feature.

\begin{table}[h]
    \centering
    \caption{Anomaly detection and localization performance of ProtoAD with different FINCH partitions. Each tuple shows image-level AUROC, pixel-level AUROC, and average cluster numbers.}
    \label{tab:finch_hierarchy}
    % \resizebox{0.8\linewidth}{!}{
    \setlength{\tabcolsep}{4.0mm}{
    \begin{tabular}{c c c c}
        \hline
        Partition & Texture & Object & All \\ \hline
           P2  &  (98.9, 97.7, 48132) & (98.2, 97.2, 92904) & (98.5, 97.3, 77980) \\
           P3  &  (98.5, 97.6, 7802) &  (97.8, 97.3, 20616) & (98.0, 97.4, 16345) \\   
           P4  &  (98.4, 97.4, 1166) &  (97.4, 97.1, 3397)	& (97.7, 97.2, 2653) \\   
           P5  &  (98.1, 97.1, 234) &   (95.2, 96.6, 902)	& (96.1, 96.8, 679) \\    
           P6  &  (95.8, 96.4, 56) &    (92.2, 95.9, 190) & (93.4, 96.1, 146) \\ \hline
         Best  &  (99.0, 97.7, 17559) & (98.2, 97.2, 52844) & (98.5, 97.4, 41083) \\ \hline
          Ours  &  (98.3, 97.5, 1787) &  (97.4, 97.1, 4626) & (97.7, 97.2, 3680) \\ \hline
    \end{tabular}}
\end{table}

\subsubsection{Partition Selection from Clustering Hierarchy}

\label{ablation:cluster_hierarchy}

FINCH is a hierarchical agglomerative clustering algorithm. It recursively merges clusters from the bottom up and provides a set of partitions in a hierarchical structure. Each successive partition is a super-set of its preceding partitions, and the number of clusters in it is smaller than those in the preceding partitions. Thus, we need select a partition from the clustering hierarchy as the clustering result.  

We report the performance of our method with different partitions, from the second (P2) to the 6-th (P6) partition of FINCH, in Table \ref{tab:finch_hierarchy} (see Table 1 in supplementary for more detailed results). We do not include the first partition because it has a huge number of clusters. The results in Table \ref{tab:finch_hierarchy} indicate the average performance decreases along with the merging process. This may be because, when the number of clusters gets smaller, clusters are less compact and unsuitable for anomaly detection. On the other hand, if the number of clusters is too large, there are too many prototypes, and the inference time and storage would increase rapidly. We also give the ``Best'' performance, which FINCH can achieve by selecting the best partition for each category respectively. This best performance is the upper bound that our method can achieve. However, selecting partition based on the average performance (from P2 to P6) or performance for each category (Best) is time-consuming and not suitable for real applications. In our method, we stop FINCH when the number of cluster is less than 10,000 and use the final partition as the clustering result, and give its results in the last line of Table \ref{tab:finch_hierarchy}. Our partition selection rule can achieve performance very close to the best one with only a tenth of clusters. Therefore, our method can reach a good trade-off between effectiveness and efficiency.

\begin{table}[h]
    \centering
    \caption{Anomaly detection and localization performance of ProtoAD with different clustering methods. Each tuple shows image-Level AUROC and pixel-level AUROC.}
    \label{tab:mvtec_kmeans_cosine}
    \setlength{\tabcolsep}{4.5mm}{
    \begin{tabular}{c c c c}
        \hline
                 &    K-Means    &   K-Means  &     FINCH  \\ 
        Category &    (L2)       &  (Norm L2) &   (Coisne) \\ \hline
        texture  &  (95.0, 95.3) & (97.2, 96.6)	& (98.3, 97.5) \\ \hline
        object	 &  (92.9, 95.9) & (95.4, 96.1)	& (97.4, 97.1) \\ \hline
        all	     &  (93.6, 95.7) & (96.0, 96.2)	& (97.7, 97.2) \\ \hline
    \end{tabular}}
\end{table}

\subsubsection{FINCH vs. K-Means}

We compare FINCH clustering algorithm with k-means for the prototype-based anomaly detection. In our method, we choose the partition generated so far by FINCH which having less than 10,000 clusters as the clustering result. For a fair comparison, we set k to 10,000 for k-means. The results in Table \ref{tab:mvtec_kmeans_cosine} indicate that the method based on FINCH (the third column) achieves better performance than that based on k-means (the first column). Although it may achieve better performance for k-means by tuning k, it is time-consuming and not feasible for real applications.

\subsubsection{Feature Normalization and Cosine Similarity}

We also explore the importance of feature normalization for the prototype-based anomaly detection. As shown in Table \ref{tab:mvtec_kmeans_cosine}, k-means with Euclidean distance on the $L2$-normalized features (Norm L2) outperforms k-means with Euclidean distance on the original features (L2) in both anomaly detection and anomaly localization and achieves greater improvements in anomaly detection. 

When the features are $L2$-normalized, cosine similarity and Euclidean distance are equivalent in the sense of nearest neighbor searching. Therefore, we use cosine similarity for clustering and measuring the deviation of test patch features from norm patch features. We further implement cosine similarity with a $1\times1$ convolution and append it to the feature extraction network. Therefore inference can be performed in an end-to-end manner.

\begin{table*}[h]
    \centering
    \caption{{Anomaly detection and localization performance on BTAD (Image-level and Pixel-level AUROC). The best results are bold-faced.}}
    \label{tab:btad}
    \resizebox{\textwidth}{!}{
    \begin{tabular}{c c c c c c c c}
        \hline
        {\bf Category} \bf & {AE(MSE)} \bf & {AE(MSE+SSIM)} \bf & {VT-ADL} \bf & {SPADE} \bf & {PaDiM} \bf & {PatchCore} \bf & {ProtoAD}  \\ \hline
        {01} & {( - , 49.0)} & {( - , 53.0)} & {( - , 99.0)} & {(93.2, 90.2)} & {(100.0, 97.2)} & {(95.4, 96.2)} & {(97.0, 95.5)} \\
        {02} & {( - , 92.0)} & {( - , 96.0)} & {( - , 94.0)} & {(74.8, 93.5)} &	{(81.5, 95.4)} &	{(85.1, 95.2)} & {(85.2, 96.5)} \\
        {03} & {( - , 95.0)} & {( - , 89.0)} & {( - , 77.0)} & {(99.4, 96.3)} &	{(98.6, 99.6)} & {(99.7, 99.5)} & {(99.8, 99.0)} \\ \hline
        {\textbf{All}} & {( - , 78.7)} & {( - , 79.3)} & {( - , 90.0)} & {(89.1, 93.3)} & {(93.4, \textbf{97.4})} & {(93.4, 97.0)} &	{(\textbf{94.0}, 97.0)} \\    \hline
    \end{tabular}}
\end{table*}

\subsection{Results on BTAD}

In Table \ref{tab:btad}, we report the results of our method on the BTAD dataset and compare them with those of the SOTA OOD-based method (SPADE, PaDiM, and ProtoAD) and the approaches adopted in \cite{mishra2021vt}. In \cite{mishra2021vt}, three reconstruction-based methods have been evaluated, auto-encoder (AE) with MSE loss, auto-encoder with MSE and SSIM loss, and Vision-Transformer-based image anomaly detection
and localization (VT-ADL). We report the image-level and pixel-level AUROC for each category and their average for all categories. For anomaly detection, ProtoAD achieved the best image-level AUROC. For anomaly localization, ProtoAD achieved the second-best pixel-level AUROC (97.0), very close to the best one (97.4) achieved by PaDiM. These results show our method’s potential to generalize to new anomalous scenarios.

\section{Conclusion}
We propose ProtoAD, a new OOD-based image anomaly detection and localization method. First, a pre-trained neural network is used to extract features for image patches. Then, a non-parametric clustering algorithm learns the prototypes for normal patch features. Finally, an image anomaly detection and localization network is constructed by appending the feature extraction network with the $L2$ feature normalization, a $1\times1$ convolutional layer, a channel max-pooling, and a subtraction operation. As a result, ProtoAD does not need a network training process and can conduct anomaly detection and localization in an end-to-end manner. Experimental results on the MVTec AD dataset and the BTAD dataset show that ProtoAD can achieve competitive performance compared to state-of-the-art methods. Furthermore, compared to other OOD-based methods, ProtoAD is more elegant and efficient. And compared to the reconstruct-based methods, ProtoAD does not need a cumbersome network training process. Therefore, it can better meet the requirements of real applications.

\section{Funding and Competing interests}
This research was supported by the National Defense Basic Scientific Research Program of China under Grant JCKY2020903B002.
The authors have no relevant financial or non-financial interests to disclose.

\bibliography{sn-bibliography}% common bib file


\begin{thebibliography}{42}
\ifx \bisbn   \undefined \def \bisbn  #1{ISBN #1}\fi
\ifx \binits  \undefined \def \binits#1{#1}\fi
\ifx \bauthor  \undefined \def \bauthor#1{#1}\fi
\ifx \batitle  \undefined \def \batitle#1{#1}\fi
\ifx \bjtitle  \undefined \def \bjtitle#1{#1}\fi
\ifx \bvolume  \undefined \def \bvolume#1{\textbf{#1}}\fi
\ifx \byear  \undefined \def \byear#1{#1}\fi
\ifx \bissue  \undefined \def \bissue#1{#1}\fi
\ifx \bfpage  \undefined \def \bfpage#1{#1}\fi
\ifx \blpage  \undefined \def \blpage #1{#1}\fi
\ifx \burl  \undefined \def \burl#1{\textsf{#1}}\fi
\ifx \doiurl  \undefined \def \doiurl#1{\url{https://doi.org/#1}}\fi
\ifx \betal  \undefined \def \betal{\textit{et al.}}\fi
\ifx \binstitute  \undefined \def \binstitute#1{#1}\fi
\ifx \binstitutionaled  \undefined \def \binstitutionaled#1{#1}\fi
\ifx \bctitle  \undefined \def \bctitle#1{#1}\fi
\ifx \beditor  \undefined \def \beditor#1{#1}\fi
\ifx \bpublisher  \undefined \def \bpublisher#1{#1}\fi
\ifx \bbtitle  \undefined \def \bbtitle#1{#1}\fi
\ifx \bedition  \undefined \def \bedition#1{#1}\fi
\ifx \bseriesno  \undefined \def \bseriesno#1{#1}\fi
\ifx \blocation  \undefined \def \blocation#1{#1}\fi
\ifx \bsertitle  \undefined \def \bsertitle#1{#1}\fi
\ifx \bsnm \undefined \def \bsnm#1{#1}\fi
\ifx \bsuffix \undefined \def \bsuffix#1{#1}\fi
\ifx \bparticle \undefined \def \bparticle#1{#1}\fi
\ifx \barticle \undefined \def \barticle#1{#1}\fi
\bibcommenthead
\ifx \bconfdate \undefined \def \bconfdate #1{#1}\fi
\ifx \botherref \undefined \def \botherref #1{#1}\fi
\ifx \url \undefined \def \url#1{\textsf{#1}}\fi
\ifx \bchapter \undefined \def \bchapter#1{#1}\fi
\ifx \bbook \undefined \def \bbook#1{#1}\fi
\ifx \bcomment \undefined \def \bcomment#1{#1}\fi
\ifx \oauthor \undefined \def \oauthor#1{#1}\fi
\ifx \citeauthoryear \undefined \def \citeauthoryear#1{#1}\fi
\ifx \endbibitem  \undefined \def \endbibitem {}\fi
\ifx \bconflocation  \undefined \def \bconflocation#1{#1}\fi
\ifx \arxivurl  \undefined \def \arxivurl#1{\textsf{#1}}\fi
\csname PreBibitemsHook\endcsname

%%% 1
\bibitem{chandola2009anomaly}
\begin{barticle}
\bauthor{\bsnm{Chandola}, \binits{V.}},
\bauthor{\bsnm{Banerjee}, \binits{A.}},
\bauthor{\bsnm{Kumar}, \binits{V.}}:
\batitle{Anomaly detection: A survey}.
\bjtitle{ACM computing surveys (CSUR)}
\bvolume{41}(\bissue{3}),
\bfpage{1}--\blpage{58}
(\byear{2009})
\end{barticle}
\endbibitem

%%% 2
\bibitem{salehi2022unified}
\begin{botherref}
\oauthor{\bsnm{Salehi}, \binits{M.}},
\oauthor{\bsnm{Mirzaei}, \binits{H.}},
\oauthor{\bsnm{Hendrycks}, \binits{D.}},
\oauthor{\bsnm{Li}, \binits{Y.}},
\oauthor{\bsnm{Rohban}, \binits{M.}},
\oauthor{\bsnm{Sabokrou}, \binits{M.}}, et al.:
A unified survey on anomaly, novelty, open-set, and out of-distribution detection: Solutions and future challenges.
Transactions on Machine Learning Research
(234)
(2022)
\end{botherref}
\endbibitem

%%% 3
\bibitem{ruff2018deep}
\begin{bchapter}
\bauthor{\bsnm{Ruff}, \binits{L.}},
\bauthor{\bsnm{Vandermeulen}, \binits{R.}},
\bauthor{\bsnm{Goernitz}, \binits{N.}},
\bauthor{\bsnm{Deecke}, \binits{L.}},
\bauthor{\bsnm{Siddiqui}, \binits{S.A.}},
\bauthor{\bsnm{Binder}, \binits{A.}},
\bauthor{\bsnm{M{\"u}ller}, \binits{E.}},
\bauthor{\bsnm{Kloft}, \binits{M.}}:
\bctitle{Deep one-class classification}.
In: \bbtitle{International Conference on Machine Learning},
pp. \bfpage{4393}--\blpage{4402}
(\byear{2018})
\end{bchapter}
\endbibitem

%%% 4
\bibitem{sabokrou2018adversarially}
\begin{bchapter}
\bauthor{\bsnm{Sabokrou}, \binits{M.}},
\bauthor{\bsnm{Khalooei}, \binits{M.}},
\bauthor{\bsnm{Fathy}, \binits{M.}},
\bauthor{\bsnm{Adeli}, \binits{E.}}:
\bctitle{Adversarially learned one-class classifier for novelty detection}.
In: \bbtitle{Proceedings of the IEEE Conference on Computer Vision and Pattern Recognition},
pp. \bfpage{3379}--\blpage{3388}
(\byear{2018})
\end{bchapter}
\endbibitem

%%% 5
\bibitem{golan2018deep}
\begin{botherref}
\oauthor{\bsnm{Golan}, \binits{I.}},
\oauthor{\bsnm{El-Yaniv}, \binits{R.}}:
Deep anomaly detection using geometric transformations.
Advances in neural information processing systems
\textbf{31}
(2018)
\end{botherref}
\endbibitem

%%% 6
\bibitem{bergman2020classification}
\begin{botherref}
\oauthor{\bsnm{Bergman}, \binits{L.}},
\oauthor{\bsnm{Hoshen}, \binits{Y.}}:
Classification-based anomaly detection for general data.
arXiv preprint arXiv:2005.02359
(2020)
\end{botherref}
\endbibitem

%%% 7
\bibitem{mei2018automatic}
\begin{barticle}
\bauthor{\bsnm{Mei}, \binits{S.}},
\bauthor{\bsnm{Wang}, \binits{Y.}},
\bauthor{\bsnm{Wen}, \binits{G.}}:
\batitle{Automatic fabric defect detection with a multi-scale convolutional denoising autoencoder network model}.
\bjtitle{Sensors}
\bvolume{18}(\bissue{4}),
\bfpage{1064}
(\byear{2018})
\end{barticle}
\endbibitem

%%% 8
\bibitem{bergmann2019mvtec}
\begin{bchapter}
\bauthor{\bsnm{Bergmann}, \binits{P.}},
\bauthor{\bsnm{Fauser}, \binits{M.}},
\bauthor{\bsnm{Sattlegger}, \binits{D.}},
\bauthor{\bsnm{Steger}, \binits{C.}}:
\bctitle{Mvtec ad--a comprehensive real-world dataset for unsupervised anomaly detection}.
In: \bbtitle{Proceedings of the IEEE/CVF Conference on Computer Vision and Pattern Recognition},
pp. \bfpage{9592}--\blpage{9600}
(\byear{2019})
\end{bchapter}
\endbibitem

%%% 9
\bibitem{sabokrou2017deep}
\begin{barticle}
\bauthor{\bsnm{Sabokrou}, \binits{M.}},
\bauthor{\bsnm{Fayyaz}, \binits{M.}},
\bauthor{\bsnm{Fathy}, \binits{M.}},
\bauthor{\bsnm{Klette}, \binits{R.}}:
\batitle{Deep-cascade: Cascading 3d deep neural networks for fast anomaly detection and localization in crowded scenes}.
\bjtitle{IEEE Transactions on Image Processing}
\bvolume{26}(\bissue{4}),
\bfpage{1992}--\blpage{2004}
(\byear{2017})
\end{barticle}
\endbibitem

%%% 10
\bibitem{sabokrou2018deep}
\begin{barticle}
\bauthor{\bsnm{Sabokrou}, \binits{M.}},
\bauthor{\bsnm{Fayyaz}, \binits{M.}},
\bauthor{\bsnm{Fathy}, \binits{M.}},
\bauthor{\bsnm{Moayed}, \binits{Z.}},
\bauthor{\bsnm{Klette}, \binits{R.}}:
\batitle{Deep-anomaly: Fully convolutional neural network for fast anomaly detection in crowded scenes}.
\bjtitle{Computer Vision and Image Understanding}
\bvolume{172},
\bfpage{88}--\blpage{97}
(\byear{2018})
\end{barticle}
\endbibitem

%%% 11
\bibitem{sabokrou2018avid}
\begin{bchapter}
\bauthor{\bsnm{Sabokrou}, \binits{M.}},
\bauthor{\bsnm{Pourreza}, \binits{M.}},
\bauthor{\bsnm{Fayyaz}, \binits{M.}},
\bauthor{\bsnm{Entezari}, \binits{R.}},
\bauthor{\bsnm{Fathy}, \binits{M.}},
\bauthor{\bsnm{Gall}, \binits{J.}},
\bauthor{\bsnm{Adeli}, \binits{E.}}:
\bctitle{Avid: Adversarial visual irregularity detection}.
In: \bbtitle{Asian Conference on Computer Vision},
pp. \bfpage{488}--\blpage{505}
(\byear{2018})
\end{bchapter}
\endbibitem

%%% 12
\bibitem{schlegl2017unsupervised}
\begin{bchapter}
\bauthor{\bsnm{Schlegl}, \binits{T.}},
\bauthor{\bsnm{Seeb{\"o}ck}, \binits{P.}},
\bauthor{\bsnm{Waldstein}, \binits{S.M.}},
\bauthor{\bsnm{Schmidt-Erfurth}, \binits{U.}},
\bauthor{\bsnm{Langs}, \binits{G.}}:
\bctitle{Unsupervised anomaly detection with generative adversarial networks to guide marker discovery}.
In: \bbtitle{International Conference on Information Processing in Medical Imaging},
pp. \bfpage{146}--\blpage{157}
(\byear{2017})
\end{bchapter}
\endbibitem

%%% 13
\bibitem{li2018thoracic}
\begin{bchapter}
\bauthor{\bsnm{Li}, \binits{Z.}},
\bauthor{\bsnm{Wang}, \binits{C.}},
\bauthor{\bsnm{Han}, \binits{M.}},
\bauthor{\bsnm{Xue}, \binits{Y.}},
\bauthor{\bsnm{Wei}, \binits{W.}},
\bauthor{\bsnm{Li}, \binits{L.-J.}},
\bauthor{\bsnm{Fei-Fei}, \binits{L.}}:
\bctitle{Thoracic disease identification and localization with limited supervision}.
In: \bbtitle{Proceedings of the IEEE Conference on Computer Vision and Pattern Recognition},
pp. \bfpage{8290}--\blpage{8299}
(\byear{2018})
\end{bchapter}
\endbibitem

%%% 14
\bibitem{an2015variational}
\begin{barticle}
\bauthor{\bsnm{An}, \binits{J.}},
\bauthor{\bsnm{Cho}, \binits{S.}}:
\batitle{Variational autoencoder based anomaly detection using reconstruction probability}.
\bjtitle{Special Lecture on IE}
\bvolume{2}(\bissue{1}),
\bfpage{1}--\blpage{18}
(\byear{2015})
\end{barticle}
\endbibitem

%%% 15
\bibitem{bergmann2018improving}
\begin{botherref}
\oauthor{\bsnm{Bergmann}, \binits{P.}},
\oauthor{\bsnm{L{\"o}we}, \binits{S.}},
\oauthor{\bsnm{Fauser}, \binits{M.}},
\oauthor{\bsnm{Sattlegger}, \binits{D.}},
\oauthor{\bsnm{Steger}, \binits{C.}}:
Improving unsupervised defect segmentation by applying structural similarity to autoencoders.
arXiv preprint arXiv:1807.02011
(2018)
\end{botherref}
\endbibitem

%%% 16
\bibitem{gong2019memorizing}
\begin{bchapter}
\bauthor{\bsnm{Gong}, \binits{D.}},
\bauthor{\bsnm{Liu}, \binits{L.}},
\bauthor{\bsnm{Le}, \binits{V.}},
\bauthor{\bsnm{Saha}, \binits{B.}},
\bauthor{\bsnm{Mansour}, \binits{M.R.}},
\bauthor{\bsnm{Venkatesh}, \binits{S.}},
\bauthor{\bsnm{Hengel}, \binits{A.v.d.}}:
\bctitle{Memorizing normality to detect anomaly: Memory-augmented deep autoencoder for unsupervised anomaly detection}.
In: \bbtitle{Proceedings of the IEEE/CVF International Conference on Computer Vision},
pp. \bfpage{1705}--\blpage{1714}
(\byear{2019})
\end{bchapter}
\endbibitem

%%% 17
\bibitem{liu2020towards}
\begin{bchapter}
\bauthor{\bsnm{Liu}, \binits{W.}},
\bauthor{\bsnm{Li}, \binits{R.}},
\bauthor{\bsnm{Zheng}, \binits{M.}},
\bauthor{\bsnm{Karanam}, \binits{S.}},
\bauthor{\bsnm{Wu}, \binits{Z.}},
\bauthor{\bsnm{Bhanu}, \binits{B.}},
\bauthor{\bsnm{Radke}, \binits{R.J.}},
\bauthor{\bsnm{Camps}, \binits{O.}}:
\bctitle{Towards visually explaining variational autoencoders}.
In: \bbtitle{Proceedings of the IEEE/CVF Conference on Computer Vision and Pattern Recognition},
pp. \bfpage{8642}--\blpage{8651}
(\byear{2020})
\end{bchapter}
\endbibitem

%%% 18
\bibitem{park2020learning}
\begin{bchapter}
\bauthor{\bsnm{Park}, \binits{H.}},
\bauthor{\bsnm{Noh}, \binits{J.}},
\bauthor{\bsnm{Ham}, \binits{B.}}:
\bctitle{Learning memory-guided normality for anomaly detection}.
In: \bbtitle{Proceedings of the IEEE/CVF Conference on Computer Vision and Pattern Recognition},
pp. \bfpage{14372}--\blpage{14381}
(\byear{2020})
\end{bchapter}
\endbibitem

%%% 19
\bibitem{zavrtanik2021reconstruction}
\begin{barticle}
\bauthor{\bsnm{Zavrtanik}, \binits{V.}},
\bauthor{\bsnm{Kristan}, \binits{M.}},
\bauthor{\bsnm{Sko{\v{c}}aj}, \binits{D.}}:
\batitle{Reconstruction by inpainting for visual anomaly detection}.
\bjtitle{Pattern Recognition}
\bvolume{112},
\bfpage{107706}
(\byear{2021})
\end{barticle}
\endbibitem

%%% 20
\bibitem{hou2021divide}
\begin{bchapter}
\bauthor{\bsnm{Hou}, \binits{J.}},
\bauthor{\bsnm{Zhang}, \binits{Y.}},
\bauthor{\bsnm{Zhong}, \binits{Q.}},
\bauthor{\bsnm{Xie}, \binits{D.}},
\bauthor{\bsnm{Pu}, \binits{S.}},
\bauthor{\bsnm{Zhou}, \binits{H.}}:
\bctitle{Divide-and-assemble: Learning block-wise memory for unsupervised anomaly detection}.
In: \bbtitle{Proceedings of the IEEE/CVF International Conference on Computer Vision},
pp. \bfpage{8791}--\blpage{8800}
(\byear{2021})
\end{bchapter}
\endbibitem

%%% 21
\bibitem{zavrtanik2021draem}
\begin{bchapter}
\bauthor{\bsnm{Zavrtanik}, \binits{V.}},
\bauthor{\bsnm{Kristan}, \binits{M.}},
\bauthor{\bsnm{Sko{\v{c}}aj}, \binits{D.}}:
\bctitle{Draem-a discriminatively trained reconstruction embedding for surface anomaly detection}.
In: \bbtitle{Proceedings of the IEEE/CVF International Conference on Computer Vision},
pp. \bfpage{8330}--\blpage{8339}
(\byear{2021})
\end{bchapter}
\endbibitem

%%% 22
\bibitem{bergmann2020uninformed}
\begin{bchapter}
\bauthor{\bsnm{Bergmann}, \binits{P.}},
\bauthor{\bsnm{Fauser}, \binits{M.}},
\bauthor{\bsnm{Sattlegger}, \binits{D.}},
\bauthor{\bsnm{Steger}, \binits{C.}}:
\bctitle{Uninformed students: Student-teacher anomaly detection with discriminative latent embeddings}.
In: \bbtitle{Proceedings of the IEEE/CVF Conference on Computer Vision and Pattern Recognition},
pp. \bfpage{4183}--\blpage{4192}
(\byear{2020})
\end{bchapter}
\endbibitem

%%% 23
\bibitem{salehi2021multiresolution}
\begin{bchapter}
\bauthor{\bsnm{Salehi}, \binits{M.}},
\bauthor{\bsnm{Sadjadi}, \binits{N.}},
\bauthor{\bsnm{Baselizadeh}, \binits{S.}},
\bauthor{\bsnm{Rohban}, \binits{M.H.}},
\bauthor{\bsnm{Rabiee}, \binits{H.R.}}:
\bctitle{Multiresolution knowledge distillation for anomaly detection}.
In: \bbtitle{Proceedings of the IEEE/CVF Conference on Computer Vision and Pattern Recognition},
pp. \bfpage{14902}--\blpage{14912}
(\byear{2021})
\end{bchapter}
\endbibitem

%%% 24
\bibitem{wang2021glancing}
\begin{bchapter}
\bauthor{\bsnm{Wang}, \binits{S.}},
\bauthor{\bsnm{Wu}, \binits{L.}},
\bauthor{\bsnm{Cui}, \binits{L.}},
\bauthor{\bsnm{Shen}, \binits{Y.}}:
\bctitle{Glancing at the patch: Anomaly localization with global and local feature comparison}.
In: \bbtitle{Proceedings of the IEEE/CVF Conference on Computer Vision and Pattern Recognition},
pp. \bfpage{254}--\blpage{263}
(\byear{2021})
\end{bchapter}
\endbibitem

%%% 25
\bibitem{yi2021patch}
\begin{bchapter}
\bauthor{\bsnm{Yi}, \binits{J.}},
\bauthor{\bsnm{Yoon}, \binits{S.}}:
\bctitle{Patch svdd: Patch-level svdd for anomaly detection and segmentation}.
In: \bbtitle{Computer Vision--ACCV 2020: 15th Asian Conference on Computer Vision, Kyoto, Japan, November 30--December 4, 2020, Revised Selected Papers, Part VI 15},
pp. \bfpage{375}--\blpage{390}
(\byear{2021})
\end{bchapter}
\endbibitem

%%% 26
\bibitem{li2021cutpaste}
\begin{bchapter}
\bauthor{\bsnm{Li}, \binits{C.-L.}},
\bauthor{\bsnm{Sohn}, \binits{K.}},
\bauthor{\bsnm{Yoon}, \binits{J.}},
\bauthor{\bsnm{Pfister}, \binits{T.}}:
\bctitle{Cutpaste: Self-supervised learning for anomaly detection and localization}.
In: \bbtitle{Proceedings of the IEEE/CVF Conference on Computer Vision and Pattern Recognition},
pp. \bfpage{9664}--\blpage{9674}
(\byear{2021})
\end{bchapter}
\endbibitem

%%% 27
\bibitem{napoletano2018anomaly}
\begin{barticle}
\bauthor{\bsnm{Napoletano}, \binits{P.}},
\bauthor{\bsnm{Piccoli}, \binits{F.}},
\bauthor{\bsnm{Schettini}, \binits{R.}}:
\batitle{Anomaly detection in nanofibrous materials by cnn-based self-similarity}.
\bjtitle{Sensors}
\bvolume{18}(\bissue{1}),
\bfpage{209}
(\byear{2018})
\end{barticle}
\endbibitem

%%% 28
\bibitem{cohen2020sub}
\begin{botherref}
\oauthor{\bsnm{Cohen}, \binits{N.}},
\oauthor{\bsnm{Hoshen}, \binits{Y.}}:
Sub-image anomaly detection with deep pyramid correspondences.
arXiv preprint arXiv:2005.02357
(2020)
\end{botherref}
\endbibitem

%%% 29
\bibitem{defard2021padim}
\begin{bchapter}
\bauthor{\bsnm{Defard}, \binits{T.}},
\bauthor{\bsnm{Setkov}, \binits{A.}},
\bauthor{\bsnm{Loesch}, \binits{A.}},
\bauthor{\bsnm{Audigier}, \binits{R.}}:
\bctitle{Padim: a patch distribution modeling framework for anomaly detection and localization}.
In: \bbtitle{International Conference on Pattern Recognition},
pp. \bfpage{475}--\blpage{489}
(\byear{2021})
\end{bchapter}
\endbibitem

%%% 30
\bibitem{roth2022towards}
\begin{bchapter}
\bauthor{\bsnm{Roth}, \binits{K.}},
\bauthor{\bsnm{Pemula}, \binits{L.}},
\bauthor{\bsnm{Zepeda}, \binits{J.}},
\bauthor{\bsnm{Sch{\"o}lkopf}, \binits{B.}},
\bauthor{\bsnm{Brox}, \binits{T.}},
\bauthor{\bsnm{Gehler}, \binits{P.}}:
\bctitle{Towards total recall in industrial anomaly detection}.
In: \bbtitle{Proceedings of the IEEE/CVF Conference on Computer Vision and Pattern Recognition},
pp. \bfpage{14318}--\blpage{14328}
(\byear{2022})
\end{bchapter}
\endbibitem

%%% 31
\bibitem{deng2009imagenet}
\begin{bchapter}
\bauthor{\bsnm{Deng}, \binits{J.}},
\bauthor{\bsnm{Dong}, \binits{W.}},
\bauthor{\bsnm{Socher}, \binits{R.}},
\bauthor{\bsnm{Li}, \binits{L.-J.}},
\bauthor{\bsnm{Li}, \binits{K.}},
\bauthor{\bsnm{Fei-Fei}, \binits{L.}}:
\bctitle{Imagenet: A large-scale hierarchical image database}.
In: \bbtitle{2009 IEEE Conference on Computer Vision and Pattern Recognition},
pp. \bfpage{248}--\blpage{255}
(\byear{2009})
\end{bchapter}
\endbibitem

%%% 32
\bibitem{mishra2021vt}
\begin{bchapter}
\bauthor{\bsnm{Mishra}, \binits{P.}},
\bauthor{\bsnm{Verk}, \binits{R.}},
\bauthor{\bsnm{Fornasier}, \binits{D.}},
\bauthor{\bsnm{Piciarelli}, \binits{C.}},
\bauthor{\bsnm{Foresti}, \binits{G.L.}}:
\bctitle{Vt-adl: A vision transformer network for image anomaly detection and localization}.
In: \bbtitle{2021 IEEE 30th International Symposium on Industrial Electronics (ISIE)},
pp. \bfpage{01}--\blpage{06}
(\byear{2021})
\end{bchapter}
\endbibitem

%%% 33
\bibitem{1965Some}
\begin{bchapter}
\bauthor{\bsnm{Macqueen}, \binits{J.}}:
\bctitle{Some methods for classification and analysis of multivariate observations}.
In: \bbtitle{Proc of Berkeley Symposium on Mathematical Statistics \& Probability}
(\byear{1965})
\end{bchapter}
\endbibitem

%%% 34
\bibitem{ester1996density}
\begin{bchapter}
\bauthor{\bsnm{Ester}, \binits{M.}},
\bauthor{\bsnm{Kriegel}, \binits{H.-P.}},
\bauthor{\bsnm{Sander}, \binits{J.}},
\bauthor{\bsnm{Xu}, \binits{X.}}, \betal:
\bctitle{A density-based algorithm for discovering clusters in large spatial databases with noise.}
In: \bbtitle{Kdd},
vol. \bseriesno{96},
pp. \bfpage{226}--\blpage{231}
(\byear{1996})
\end{bchapter}
\endbibitem

%%% 35
\bibitem{von2007tutorial}
\begin{barticle}
\bauthor{\bsnm{Von~Luxburg}, \binits{U.}}:
\batitle{A tutorial on spectral clustering}.
\bjtitle{Statistics and computing}
\bvolume{17}(\bissue{4}),
\bfpage{395}--\blpage{416}
(\byear{2007})
\end{barticle}
\endbibitem

%%% 36
\bibitem{karypis1999chameleon}
\begin{barticle}
\bauthor{\bsnm{Karypis}, \binits{G.}},
\bauthor{\bsnm{Han}, \binits{E.-H.}},
\bauthor{\bsnm{Kumar}, \binits{V.}}:
\batitle{Chameleon: Hierarchical clustering using dynamic modeling}.
\bjtitle{Computer}
\bvolume{32}(\bissue{8}),
\bfpage{68}--\blpage{75}
(\byear{1999})
\end{barticle}
\endbibitem

%%% 37
\bibitem{sarfraz2019efficient}
\begin{bchapter}
\bauthor{\bsnm{Sarfraz}, \binits{S.}},
\bauthor{\bsnm{Sharma}, \binits{V.}},
\bauthor{\bsnm{Stiefelhagen}, \binits{R.}}:
\bctitle{Efficient parameter-free clustering using first neighbor relations}.
In: \bbtitle{Proceedings of the IEEE/CVF Conference on Computer Vision and Pattern Recognition},
pp. \bfpage{8934}--\blpage{8943}
(\byear{2019})
\end{bchapter}
\endbibitem

%%% 38
\bibitem{zagoruyko2016wide}
\begin{bchapter}
\bauthor{\bsnm{Zagoruyko}, \binits{S.}},
\bauthor{\bsnm{Komodakis}, \binits{N.}}:
\bctitle{Wide residual networks}.
In: \bbtitle{Procedings of the British Machine Vision Conference 2016}
(\byear{2016})
\end{bchapter}
\endbibitem

%%% 39
\bibitem{he2016deep}
\begin{bchapter}
\bauthor{\bsnm{He}, \binits{K.}},
\bauthor{\bsnm{Zhang}, \binits{X.}},
\bauthor{\bsnm{Ren}, \binits{S.}},
\bauthor{\bsnm{Sun}, \binits{J.}}:
\bctitle{Deep residual learning for image recognition}.
In: \bbtitle{Proceedings of the IEEE Conference on Computer Vision and Pattern Recognition},
pp. \bfpage{770}--\blpage{778}
(\byear{2016})
\end{bchapter}
\endbibitem

%%% 40
\bibitem{van2009python}
\begin{barticle}
\bauthor{\bsnm{Van}, \binits{R.G.}},
\bauthor{\bsnm{Drake}, \binits{F.}}:
\batitle{Python 3 reference manual}.
\bjtitle{Scotts Valley, CA: CreateSpace}
\bvolume{10},
\bfpage{1593511}
(\byear{2009})
\end{barticle}
\endbibitem

%%% 41
\bibitem{paszke2019pytorch}
\begin{botherref}
\oauthor{\bsnm{Paszke}, \binits{A.}},
\oauthor{\bsnm{Gross}, \binits{S.}},
\oauthor{\bsnm{Massa}, \binits{F.}},
\oauthor{\bsnm{Lerer}, \binits{A.}},
\oauthor{\bsnm{Bradbury}, \binits{J.}},
\oauthor{\bsnm{Chanan}, \binits{G.}},
\oauthor{\bsnm{Killeen}, \binits{T.}},
\oauthor{\bsnm{Lin}, \binits{Z.}},
\oauthor{\bsnm{Gimelshein}, \binits{N.}},
\oauthor{\bsnm{Antiga}, \binits{L.}}, et al.:
Pytorch: An imperative style, high-performance deep learning library.
Advances in neural information processing systems
\textbf{32}
(2019)
\end{botherref}
\endbibitem

%%% 42
\bibitem{johnson2019billion}
\begin{barticle}
\bauthor{\bsnm{Johnson}, \binits{J.}},
\bauthor{\bsnm{Douze}, \binits{M.}},
\bauthor{\bsnm{J{\'e}gou}, \binits{H.}}:
\batitle{Billion-scale similarity search with gpus}.
\bjtitle{IEEE Transactions on Big Data}
\bvolume{7}(\bissue{3}),
\bfpage{535}--\blpage{547}
(\byear{2019})
\end{barticle}
\endbibitem

\end{thebibliography}

\end{document}